# An Overview and Evaluation of Various Face and Eyes Detection Algorithms for Driver Fatigue Monitoring Systems

Markan Lopar and Slobodan Ribarić
Department of electronics, microelectronics, computer and intelligent systems
Faculty of Electrical Engineering and Computing, University of Zagreb
Zagreb, Croatia
markan.lopar@fer.hr, slobodan.ribaric@fer.hr

*Abstract*—In this work various methods and algorithms for face and eyes detection are examined in order to decide which of them are applicable for use in a driver fatigue monitoring system. In the case of face detection the standard Viola-Jones face detector has shown best results, while the method of finding the eye centers by means of gradients has proven to be most appropriate in the case of eyes detection. The later method has also a potential for retrieving behavioral parameters needed for estimation of the level of driver fatigue. This possibility will be examined in future work.

## I. Introduction

The driver's loss of attention due to drowsiness or fatigue is one of the major contributors in road accidents. According to National Highway Traffic Safety Administration [1] more than 100,000 crashes per year in United States are caused by drowsy driving. This is a conservative estimation, because the drowsy driving is underreported as a cause of crashes. Also the accidents caused by driver's inattention are not accounted. According to some French statistics for 2003 [2], more accidents were caused by inattention due to fatigue or sleep deprivation than by alcohol or drug intoxication. Therefore, significant efforts were made over the past two decades to develop robust and reliable safety systems intended to reduce the number of accidents, as well as to introduce new, or improve existing techniques and technologies for development of these systems.

Different techniques are used in driver-fatigue monitoring systems. These techniques are divided into four categories [3]. The first category includes intrusive techniques, which are mostly based on monitoring biomedical signals, and therefore require physical contact with the driver. The second category includes non-intrusive techniques based on visual assessment of driver's bio-behavior from face images. The third category includes methods based on driver's performance, which monitor vehicle behaviors such as moving course, steering angle, speed, braking, etc. Finally, the fourth category combines techniques from the abovementioned three categories.

The computer vision based techniques from the second category are particularly effective, because the drowsiness can be detected by observing the facial features and visual bio-behavior such as head position, gaze, eye openness, eyelid movements, and mouth openness. This category encompasses a wide range of methods and algorithms with respect to data acquisition (standard, IR and stereo cameras), image processing and feature extraction (Gabor wavelets, Gaussian derivatives, Haar-like masks, etc.), and classification (neural networks, support vector machines, knowledge-based systems, etc.). A widely accepted parameter that is measured by most of methods is percentage of eye closure over time [4], commonly referred as PERCLOS. This measure was established by in [5] as the proportion of time in a specific time interval during which the eyes are at least 80% closed, not including eye blinks. Also, there are other parameters that can be measured such as average eye closure speed (AECS) proposed by Ji and Yang [6].

The initial task in development of driver fatigue monitoring system is to build a module for reliable real-time detection and tracking of facial features such as face, eyes, mouth, gaze, and head movements. This paper examines some of existing techniques for face and eyes detection.

## II. Related Works

Many efforts in about past twenty years were made to develop feature detection systems for purpose of driver fatigue monitoring. One of widely popular methods is to use active illumination for eye detection. Grace et al. [7] have used special camera for capturing images of a driver at different wavelengths. Using the fact that at different wavelengths human retina reflects different amount of light, they subtract two images of a driver taken in this way in order to obtain an image which contains non-zero pixels at the eyes locations. Many variations of this image subtracting technique are implemented later. One such notable implementation is presented by Ji and Yang [6]. In their work they use active infra-red illumination system which consists of CCD camera with two sets of infra-red LEDs distributed evenly and

This work is supported by EU-funded IPA project "VISTA – Computer Vision Innovations for Safe Traffic".

symmetrically along the circumference of two coplanar and concentric rings, where the center of both rings coincides with the camera optical axis. The light from two rings enters the eye at different angle resulting in images with bright pupil effect in case when the inner ring is turned on, and with dark pupil effect when the outer ring is switched on. The resulting images are subtracted in order to obtain the eyes positions. This method of eye detection is used by many of other researchers [8, 9].

While the abovementioned approaches use active illumination and as such rely on special equipment, there are also methods that employ other strategies. Thus Smith, Shah, and da Vitoria Lobo [10] present a system which relies on estimation of global motion and color statistics to track a person's head and facial features. For the eyes and lips detection they employ color predicates, histogram-like structures where the cells are labeled as either a members or non-members of desired color range [11]. Another work that uses color information for facial features detection is presented by Rong-ben, Ke-you, Shu-ming, and Jiang-wei [12]. Their method is based on the assumption that in RGB color space red and green components of human skin follow Gaussian planar distribution. The R and G values of each pixel are examined, and if they fall within 3 standard deviations of R and G's means, the pixel is considered to belong to skin. When the face is localized the Gabor wavelets are applied for extracting the eye features. The same procedure is later applied for mouth detection [13].

D'Orazio, Leo, Spagnolo, and Guaragnella [14] present the method for eyes detection using operator based on Circle Hough Transform [15]. This operator is applied on the entire image and the result is a maximum that represents the region possibly containing an eye. The second eye is then searched in two opposite directions compatible with the range of possible eyes position concerning the distance and orientation between the eyes. The obtained results are subjected to testing of similarity, which is evaluated by calculating the mean absolute error applied on mirrored domains. If this similarity measure falls below certain threshold, the regions are considered the best match for eye candidate.

Ribarić, Lovrenčić, and Pavešič [3] have presented a driver fatigue monitoring system which combines a neural-network-based feature extraction module with knowledge-based inference module. The feature extraction module consists of several components that perform various tasks such as face detection, detection of in-plane and out-of-plane head rotations, and estimation of eye and mouth openness. The face detection procedure uses combined appearance-based and neural network approaches. The first step of algorithm applies a hybrid method based on some features of the HMAX model [16] and Viola-Jones face detector [17]: A linear template matching is performed using fixed-size Haar-like masks, and then a non-linear MAX operator is applied in order to gain invariance to transformations of an input image. The features obtained by applying MAX operator are used as inputs to the 3-layered perceptron in the second step of face detection algorithm. These features are chosen from fixed-size window that slides through the maps of features, and the value of a single output neuron for each position of the sliding window is recorded in an activation map. The activation map is binarized afterwards, and the center of gravity of largest region with non-zero value is considered to be the center of detected face. The estimation of the angle of the in-plane rotation is implemented using 5-layered convolutional neural network. The inputs are the features obtained in the face detection routine before applying MAX operator. The output layer has 36 neurons which correspond to various rotation angles. The estimation of the angle of the out-of-plane rotation is carried out in a similar manner, except that the angles of pan and tilt rotations are estimated from horizontal and vertical offsets from the center of the face and the point that lies in the intersection of the nose symmetry axis and the line connecting the eyes. The location of the eyes and mouth is determined on the basis of the center of the face, as well as the angles of in-plane and out-of-plane rotations. Two separate convolutional neural networks are employed to determine the eye and mouth openness.

Two more approaches are presented here which deal with the problem of eyes detection. Though they are originally not discussed in the context of driver fatigue monitoring, they can nevertheless be applied for that purpose. The first work is by Valenti and Gevers [18]. Their approach is based on the observation that eyes are characterized by radially symmetric brightness pattern, and isophotes are used to infer the center of circular patterns. The center is obtained by voting mechanism which is used to increase and weight important votes to reinforce center estimates. The second method is proposed by Timm and Barth [19]. It is relatively simple method which is based on fact that all circle gradients intersect in the center of a circle. Thus the eye centers can be obtained at locations where most of the image gradients intersect. Moreover, this method is robust enough to detect partially occluded circular objects, e.g. half-open eyes. This is great advantage over the Circle Hough Transform which in this case would fail.

III. ANALYZING OF VARIOUS METHODS FOR FACE AND EYES DETECTION

The aim of this work is to test and compare some methods for face and eyes detection and to decide which of them are feasible for implementation in driver fatigue monitoring system. The particular attention has been paid to accuracy in feature detection and ability to operate in real time. The selected methods are tested on video sequences captured with web camera in dark and light ambient and with various background scenery. The tests are performed on the machine with two cores, each working at 2,4 GHz. Since the extensive quantitative measurements are not available at the moment, the obtained results are presented in descriptive form only. These results are acquired by visual inspection: the accuracy is estimated by observing the frequency of correctly detected features, and feasibility for real time processing is evaluated with regard to how smoothly the respective algorithm runs.

A. Face Detection

Face detectors are implemented using several methods. These methods include face detection by means of searching for ellipses, detection using SIFT [20] and SURF features [21], application of Viola-Jones face detector [17], and use of feature extraction module described in [3].

Since the human head is shaped close to an ellipse, it is convenient to try to detect it by searching for ellipses in a given image. Ellipses can be found by Generalized Hough Transform [22], but this procedure is computationally intensive. Prakash and Rajesh [23] are recently presented the method for ellipse detection which reduces the whole issue to application of Circle Hough Transform, and this method is used in this work. Nevertheless, it is not fast enough to operate in real time. Also, since there are lots of false positives, it is necessary to implement some learning procedure, and this additionally slows down the detection procedure.

SIFT and SURF are well-known algorithms for object matching. Both of these methods perform well in terms of matching of facial features in template and target image, with SURF running slightly faster. Unfortunately, both of them have one major drawback: they are very sensitive to even small changes in illuminating conditions. An attempt was made to circumvent this problem by applying histogram equalization and histogram fitting [24], but with little success, since in that case the detection runs much slower.

The Viola-Jones detector is well-known face detector that has found its use in many applications that involve face detection. It performs very well considering both accuracy and speed needed for real-time processing. The facial feature extractor described in [3] which implements some features of Viola-Jones detector, also performs the task of face detection very well.

The advantages and drawbacks of each face detector are summarized in TABLE I.

TABLE I. ADVANTAGES AND DISADVANTAGES OF VARIOUS FACE DETECTION METHODS

| Detection method | Advantages | Disadvantages |
| --- | --- | --- |
| Ellipse detection | Matches well the shape of human head | Very slow, high number of false positives |
| SIFT | An excellent method for object matching | Very high sensitivity to changes in illumination conditions |
| SURF | An excellent method for object matching, faster than SIFT | Very high sensitivity to changes in illumination conditions |
| Viola-Jones | Accurate and very fast | Small number of false positives, needs learning |
| Ribarić, Lovrenčić, Pavešič [3] | Accurate and fast | Small number of false positives, long training procedure |

B. Eyes Detection

Four methods are used for the eyes detection. These include Viola-Jones eyes detector, the algorithm described by Valenti and Gevers [18], the approach proposed by Timm and Barth [19], and feature extractor by Ribarić, Lovrenčić, and Pavešič [3]. All of these methods rely on previously detected face and for purpose of face detection the Viola-Jones detector is used, since it performs the best. The exception is the method by Ribarić, Lovrenčić, and Pavešič, since it uses its own face detector.

The Viola-Jones performs reasonably well, but it produces a slightly higher amount of false positives, especially in the regions of mouth corners.

The detector described by Valenti and Gevers performs excellent regarding both speed and accuracy. It never fails to find the eyes, but it is not perfect in accurate localization of eye centers, since it more often than not misdetects the eye corner as the eye center.

The detector by Timm and Barth performs the best of all four eye detectors. It runs slightly slower than the detector by Valenti and Gevers, but it almost never fails to find accurately the eye center.

Finally, the feature extractor by Ribarić, Lovrenčić, and Pavešič is not as good in the task of eye detection as in the case of face detection. The cause of this lies in the dependency chain of detected features. The detected face is prerequisite for in-plane rotation detection. The out-of-plane rotation detection needs previously detected in-plane rotation, and finally the eyes and mouth detection takes place after the out-of-plane rotation is detected. Thus the error detection grows larger in each step of this cascade, and it has highest value at the end of the chain, i.e. in case of eyes and mouth detection.

The advantages and drawbacks of each eyes detector are summarized in TABLE II.

TABLE II. ADVANTAGES AND DISADVANTAGES OF VARIOUS EYES DETECTION METHODS

| Detection method | Advantages | Disadvantages |
| --- | --- | --- |
| Viola-Jones | Accurate and very fast | Has more false positives than the face detector, needs learning |
| Valenti and Gevers | Reliable and very fast | Misdetects the eye corners as the eye centers |
| Timm and Barth | Very accurate in eye center detection | Slower than detector by Valenti and Gevers |
| Ribarić, Lovrenčić, Pavešič [3] | Fast detector | Lower accuracy due to features dependency chain |

IV. CONCLUSIONS AND FUTURE WORK

In this work we have tested some methods for face and eyes detection in order to decide which of them is the best for implementation in driver fatigue monitoring system. Viola-Jones face detector has proven to be fastest and most accurate among the face detectors, while the algorithm proposed by Timm and Barth outperformed the rest of eyes detectors.

The eyes detection method described by Timm and Barth has also one interesting property that may be utilized in driver fatigue monitoring systems. Namely, by simple counting the number of intersecting gradients and comparing it to some referent value it is possible to determine the degree of eyes openness/closeness and to calculate some behavioral parameters such as PERCLOS. This possibility will definitely be examined in the future.

The future work also includes researching and development of methods for mouth detection and determination of degree of mouth openness. The use of thermovision camera for image acquisition is also planned in the future in order to examine whether some facial features can be obtained from images acquired in this way. Thermovision cameras are at the moment too expensive to be used in commercial driver fatigue monitoring systems, but their potential usefulness nevertheless may and should be examined in researching activities.

After the facial features are obtained, a certain number of behavioral parameters – such as PERCLOS, degree of mouth openness, degree of head rotation, and gaze estimation – will be extracted from facial features. These parameters will be used in inference module which should decide whether the driver is fatigued or not, and issue an appropriate warning or alarm signal if it is necessary. In any case, the task of monitoring driver fatigue is a challenge to be handled, and many efforts should be made to deal with it successfully.